\pdfoutput=1

\documentclass[11pt]{article}

\usepackage[final]{acl}

\usepackage{times}
\usepackage{latexsym}

\usepackage[T1]{fontenc}

\usepackage[utf8]{inputenc}

\usepackage{microtype}

\usepackage{inconsolata}

\usepackage{graphicx}
\usepackage{algorithm}
\usepackage{algorithmic}
\usepackage{makecell}
\usepackage{multirow}
\usepackage{longtable}
\usepackage{booktabs}
\usepackage{soul}
\usepackage{color, colortbl}
\definecolor{Gray}{gray}{0.9}
\usepackage{xcolor} 
\usepackage{amsmath}
\usepackage{todonotes}
\usepackage{changes}
\usepackage{enumitem}
\usepackage{fancybox}
\usepackage{tcolorbox}
\usepackage{hyperref}
\newcommand{\up}[1]{\textcolor{brown}{{$\uparrow$ #1}}}
\newcommand{\down}[1]{\textcolor{brown}{{$\downarrow$ #1}}}
\title{Token Prepending: A Training-Free Approach for Eliciting Better\\ Sentence Embeddings from LLMs}

\author{%
\textbf{Yuchen Fu}\footnotemark[1] \quad \textbf{Zifeng Cheng}\footnotemark[1] \quad \textbf{Zhiwei Jiang}\footnotemark[2] \quad \textbf{Zhonghui Wang} \\
\textbf{Yafeng Yin} \quad  \textbf{Zhengliang Li} \quad \textbf{Qing Gu}\\
State Key Laboratory for Novel Software Technology, Nanjing University, China \\
\texttt{\{yuchenfu,chengzf\}@smail.nju.edu.cn, jzw@nju.edu.cn} \\
\texttt{zhonghuiwang@smail.nju.edu.cn, yafeng@nju.edu.cn } \\
\texttt{lzl@smail.nju.edu.cn, guq@nju.edu.cn} 
}
\renewcommand{\thefootnote}{\fnsymbol{footnote}}

\begin{document}

\maketitle
\begin{abstract}
Extracting sentence embeddings from large language models (LLMs) is a promising direction, as LLMs have demonstrated stronger semantic understanding capabilities. 
Previous studies typically focus on prompt engineering to elicit sentence embeddings from LLMs by prompting the model to encode sentence information into the embedding of the last token.
However, LLMs are mostly decoder-only models with causal attention and the earlier tokens in the sentence cannot attend to the latter tokens, resulting in biased encoding of sentence information and cascading effects on the final decoded token.
To this end, we propose a novel Token Prepending (TP) technique that prepends each layer's decoded sentence embedding to the beginning of the sentence in the next layer's input, allowing earlier tokens to attend to the complete sentence information under the causal attention mechanism.
The proposed TP technique is a plug-and-play and training-free technique, which means it can be seamlessly integrated with various prompt-based sentence embedding methods and autoregressive LLMs.
Extensive experiments on various Semantic Textual Similarity (STS) tasks and downstream classification tasks demonstrate that our proposed TP technique can significantly improve the performance of existing prompt-based sentence embedding methods across different LLMs, while incurring negligible additional inference cost.
The code are available on \url{https://github.com/fuyuchenIfyw/token_prepending.git}.
\end{abstract}

\section{Introduction}

Sentence embeddings have a wide range of applications in real-world scenarios, such as information retrieval, recommender systems, sentiment analysis, document clustering, and so on.
Recently, with the success of large language models (LLMs) in zero-shot settings for various natural language processing (NLP) tasks, some researchers have begun to focus on directly extracting sentence embeddings from LLMs without the need for additional fine-tuning~\cite{liumeaning,lei2024meta}.
This training-free setup is both practical and promising, as it does not require training data, avoids the costs of fine-tuning a large-scale model, and prevents the potential loss of general semantic understanding capabilities caused by fine-tuning on specific data.

\begin{figure}[t]
    \centering
    \includegraphics[width=0.47\textwidth]{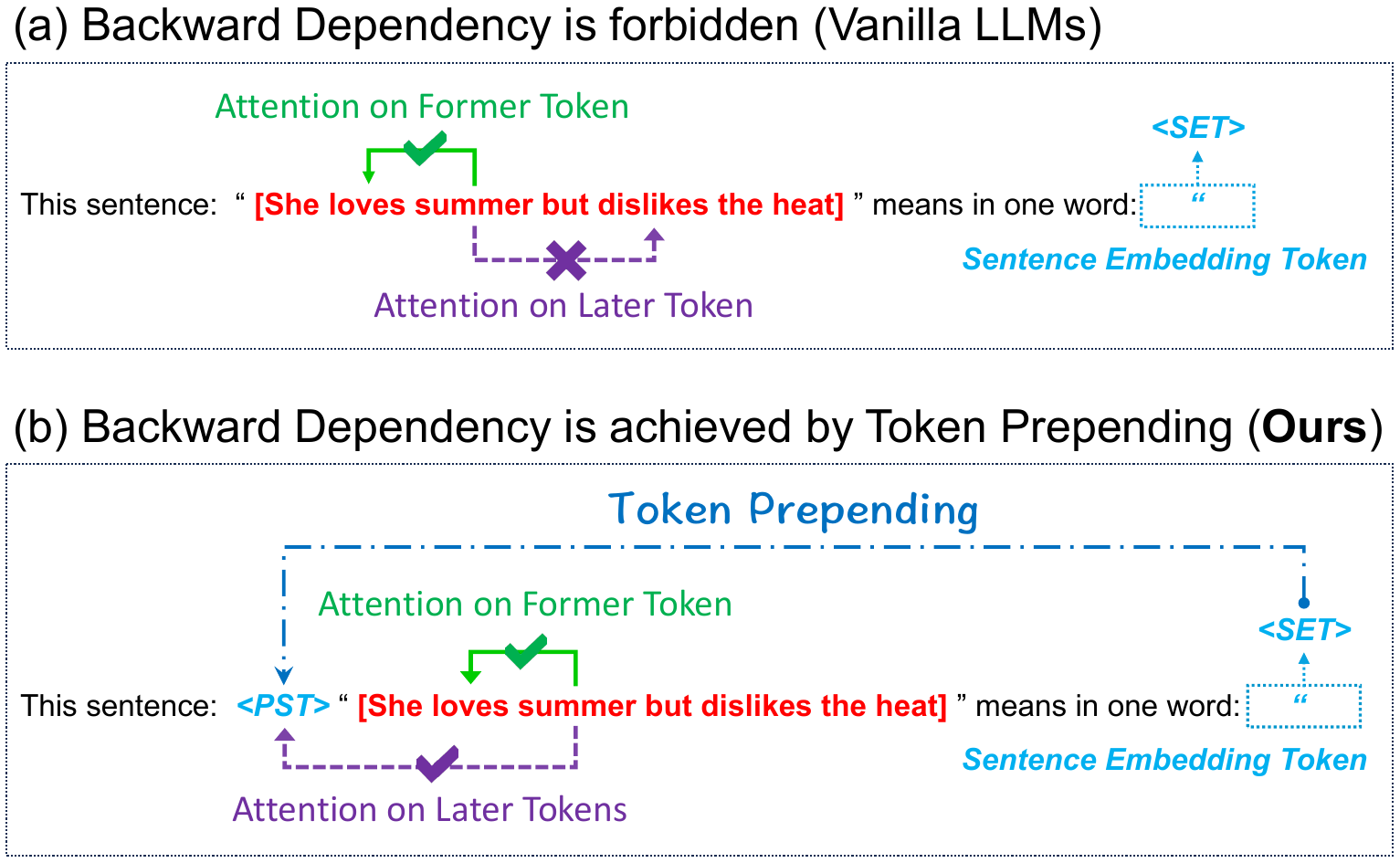}
    \caption{Comparison between (a) vanilla LLMs and (b) our proposed LLMs with token prepending.}
    \label{fig:concept}
\end{figure}

Different from previous encoder-only bidirectional langauge model like BERT~\cite{DBLP:conf/naacl/DevlinCLT19}, current LLMs are mostly decoder-only models with causal attention~\cite{touvron2023llama,brown2020language}, which make the earlier tokens in the sentence cannot attend to the latter tokens, as shown in Figure~\ref{fig:concept}(a).
To this end, recent studies~\cite{jiang2023scaling,lei2024meta,zhang2024simple} attempt to prompt the model to encode sentence information into the embedding of the last token (i.e., the {\textless \text{SET}\textgreater} in Figure~\ref{fig:concept}(a)), which can attend to all preceding tokens, thereby avoiding the problem of backward dependency.
Among the prompt-based methods,~\citet{jiang2023scaling} first propose to use a simple and effective prompt (e.g., the prompt in Figure~\ref{fig:concept}(a)) to extract sentence embeddings from LLMs.
Later, meta-task prompts~\cite{lei2024meta} and prompts with CoT and Knowledge Enhancement~\cite{zhang2024simple} are employed to extract sentence embeddings.

However, even if the last token is able to attend to all tokens in the sentence under the causal attention mechanism, the earlier tokens in the sentence still cannot attend to the later tokens (i.e., the backward dependency in Figure~\ref{fig:concept}). 
This results in biased encoding of sentence information and cascading effects on the last token.
To address this problem, some previous work \cite{springer2024repetition} has attempted to achieve backward dependency through repetition. 
They point out that processing the input twice allows LLMs to have a deeper understanding of the sentence and improves performance on various tasks. 
Nonetheless, repetition significantly increases the sequence length and substantially alters the sentence structure, leading to higher inference costs and less ideal performance.

In this paper, we propose a simple yet effective technique called Token Prepending (TP). 
As shown in Figure~\ref{fig:concept}(b), our core idea is to prepends each layer's decoded sentence embedding to the beginning of the sentence in the next layer's input, allowing earlier tokens to attend to the complete sentence information under the causal attention mechanism.
Notably, the TP technique is entirely training-free, as it introduces no additional learnable parameters.
Specifically, although TP is applicable to all layers, we find that it is not necessary to perform the TP operation across all layers. 
Instead, performing this operation only in the early layers of the model yields better performance. 
Therefore, we discontinue the TP operation after several early layers and revert to standard forward propagation. 
Additionally, considering that the final layer of LLMs is primarily used for token generation and contains less semantic information \cite{liu2024fantastic,jin2024exploring}, we propose an early-exit strategy that outputs embeddings from intermediate layers, rather than the final layer, to serve as sentence embeddings.

Our main contributions are as follows:
\begin{itemize}
\vspace{-0.4em}
\item We propose a novel TP technique for eliciting sentence embeddings from LLMs. This plug-and-play technique neither introduces new parameters nor alters the existing ones, allowing it to be seamlessly integrated with various prompt-based sentence embedding methods and autoregressive LLMs. Moreover, it adds only a single token to the original sentence, resulting in minimal additional inference overhead.
\vspace{-0.4em}
\item We perform an in-depth exploration of the TP technique and identify the most effective ways to utilize it, including the optimal layer scope of operation and the early exit strategy.
\item We conduct extensive experiments on various Semantic Textual Similarity (STS) benchmarks and downstream classification tasks. The results demonstrate that our proposed TP technique can significantly improve the performance of existing prompt-based sentence embedding methods across different LLMs.
\end{itemize}

\section{Related Work}

\noindent \textbf{Sentence Embeddings}\indent 
Sentence embedding is a fundamental task in natural language processing, aiming to map the semantic information of sentences into fixed-size vector representations.
Previous research often employs unsupervised or supervised contrastive learning to fine-tune smaller pre-trained models to enhance sentence embeddings~\cite{gao2021simcse,DBLP:conf/emnlp/JiangJHZWZWHDZ22,DBLP:conf/acl/NiACMHCY22,DBLP:conf/acl/ChanchaniH23,DBLP:conf/acl/SuSKWHOYSZ023}.
For example, Sentence-T5~\cite{DBLP:conf/acl/NiACMHCY22} explores three strategies to extract T5~\cite{DBLP:journals/jmlr/RaffelSRLNMZLL20} sentence representations and uses two-stage training to refine T5 sentence embeddings.
Unlike these methods, we focus on sentence embeddings extracted by large language models without the need for fine-tuning.

\noindent \textbf{LLMs for Sentence Embeddings}\indent
Recently, a series of studies focus on enhancing the sentence embedding of LLMs with causal attention mechanism through fine-tuning \cite{li2024bellm,behnamghader2024llm2vec,NV-Embed/Lee,generative/muennighoff}.
Due to the limited representation learning capability of unidirectional attention in LLMs, these methods mostly replace it with bidirectional attention and fine-tune LLMs using contrastive learning.
For example, BeLLM \cite{li2024bellm} converts the last attention layer from unidirectional to bidirectional and uses SimCSE~\cite{gao2021simcse} to fine-tune LLMs.
However, fine-tuning LLMs is very expensive and inevitably results in the loss of their other general capabilities.
Thus, this paper focuses on extracting sentence embeddings from LLMs without fine-tuning.

\noindent \textbf{Extracting Sentence Embeddings from LLMs}\indent 
Existing methods on extracting sentence embeddings from LLMs mainly focus on designing prompts to improve sentence embeddings.
PromptEOL \cite{jiang2023scaling} demonstrates the potential of LLMs in generating sentence embeddings by leveraging prompt engineering.
Echo embeddings \cite{springer2024repetition} repeats the input twice within the context and extracts embeddings from the second occurrence, allowing early token embeddings to encode information about subsequent tokens.
MetaEOL \cite{lei2024meta} designs meta-task prompts via ChatGPT-4 to guide LLMs to consider sentence representations from multiple perspectives.
Pretended CoT \cite{zhang2024simple} uses CoT to inspire the model to output better embeddings.
Knowledge Enhancement \cite{zhang2024simple} provides explicit guidance to the model by conveying human experience in text summarization through prompts. 
CP \cite{cheng2025contrastive} introduces an extra auxiliary prompt to elicit better sentence embedding.
In this paper, we propose a plug-and-play technique TP to improve the various prompt-based methods with negligible additional inference cost.

\begin{figure*}[t]
    \centering
    \includegraphics[width=0.97\textwidth]{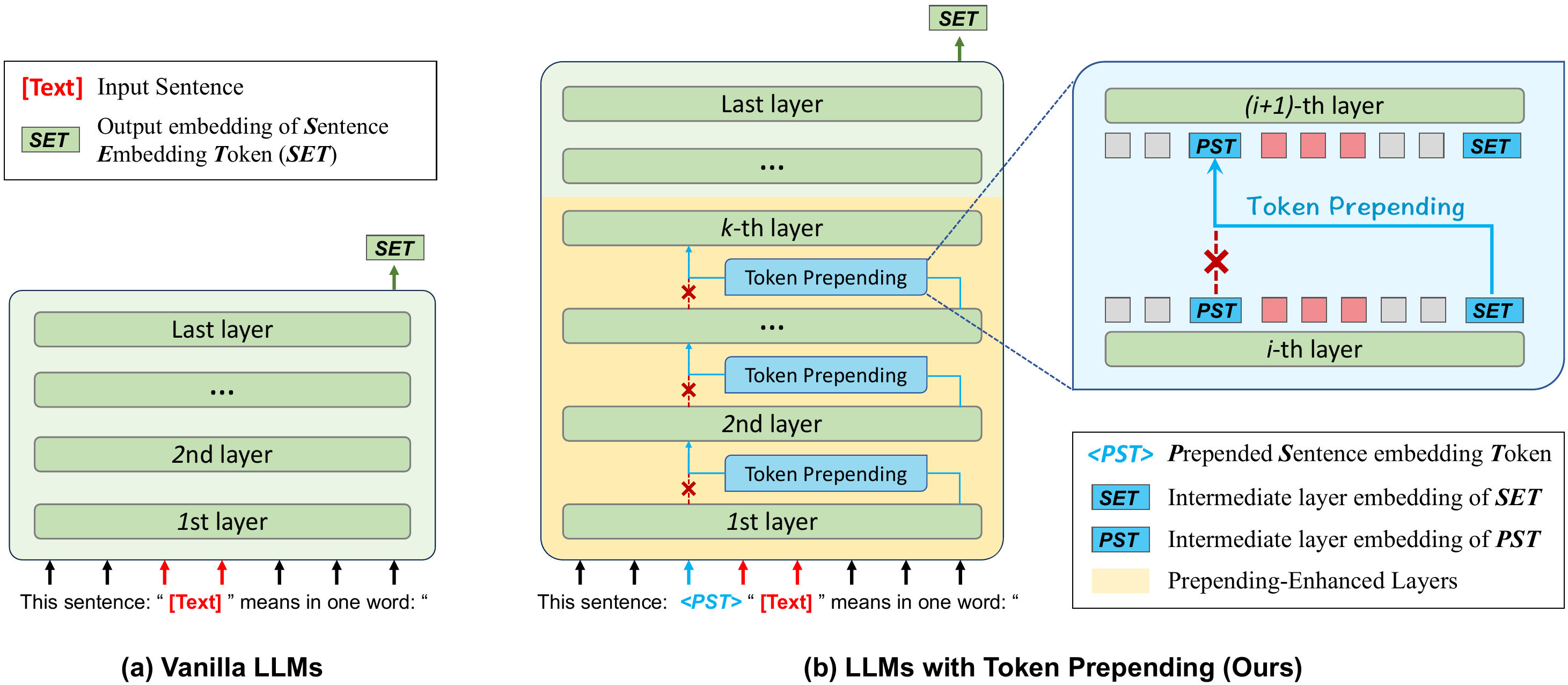}
    \caption{Illustration of extracting sentence embeddings from (a) vanilla LLMs and (b) LLMs with Token Prepending.}
    \label{fig:framework}
\end{figure*}

\section{Preliminary}

Previous work mainly focused on eliciting sentence embeddings from LLMs through prompt engineering. 
This process does not interfere with the internal operations of the LLMs but simply guides their behavior through different prompts.
As shown in Figure \ref{fig:framework}(a), PromptEOL~\cite{jiang2023scaling} introduces a widely adopted template for extracting sentence embeddings from LLMs:
\begin{equation}
    \text{This sentence: ``[Text]'' means in one word: ``} \nonumber
\end{equation}
where [Text] denotes the placeholder for the input sentence and the last token `` is used to decode the Sentence Embedding Token (SET).
The phrase ``in one word'' is a constraint that can prevent LLMs from generating long sentences, limiting a sentence to being represented by the embedding of a single word.

Formally, given the input $T = [t_1, ..., t_n]$ wrapped in a template, we first obtain the embeddings $\textbf{h}^0 = [h^0_1, \cdots, h^0_n]$ through the embedding layer, and then pass them into the $L$ Transformer layers of LLMs.
The previous work (e.g., PromptEOL) use the last layer's hidden state for the sentence embedding token $\textbf{h}^L_n$ as the output sentence embedding.
Specifically,
\begin{equation}
    \textbf{h}^{L} = {\rm LLM}^{1:L}(\textbf{h}^{0}) \nonumber
\end{equation}

\section{Proposed Method}

\subsection{Overview}
Different from previous work that only focuses on prompt engineering, our proposed method slightly intervenes in the internal operations of the LLMs.
Our core idea is to prepend the decoded sentence embedding token from the previous layer to the sentence in the next layer's input, making the semantics of the sentence perceptible to all tokens in the target sentence.
As shown in Figure \ref{fig:framework}(b), we perform the token prepending (TP) operation within the layer scope of the first few layers, which is denoted in yellow.
For the input layer, we prepend a special {\textless \text{PST}\textgreater}  token to the input sentence (i.e., [Text]) in prompt.
For the intermediate layers, we perform the TP operation between two layers by replacing the embedding of the {\textless \text{PST}\textgreater} token with the sentence embedding decoded from the last token of prompt. 
By repeating this operation across several layers, the embedding of the {\textless \text{PST}\textgreater} token may contain sufficient sentence information, or all tokens of the target sentence may perceive enough sentence information.
After that, we will discontinue the TP operation.
Finally, considering that the last layer of LLMs is primarily used for token generation, we will choose a sentence embedding from an intermediate layer as the output sentence embedding.

\subsection{Token Prepending}
Our proposed TP technique is a plug-and-play operation primarily used to adjust context dependency by intervening in the inputs of LLM layers. From the perspective of its operating layer, it can be described in detail from the following three aspects.

\subsubsection{Initial Token Prepending}
We first conduct initial token prepending operation that prepends the sentence embedding token to the input text as shown in Figure 2(b).
Since the sentence embedding token is not available at this stage, we prepend a custom token ``{\textless \text{PST}\textgreater}", which is not included in the LLM's vocabulary, serving as a placeholder for sentence embedding token.
We randomly initialize the parameters of this token and incorporate it into the input for the first Transformer layer.
Consequently, the modified embedding layer output is denoted as $\textbf{h}^0 = [h^0_1, \cdots, h_{i-1}^{0}, h_{i^*}^0, h_{i}^{0}, \cdots, h^0_n]$, where \( h_{i^*}^0 \) represents the initialized embedding of the {\textless \text{PST}\textgreater} token.

\subsubsection{Intermediate Token Prepending}
After initial token prepending, the input passes through the pretending-enhanced layers, where each layer consists of a standard Transformer layer and a specially designed intermediate token prepending.
For intermediate token prepending, we pretend the sentence embedding token {\textless \text{SET}\textgreater} to replace {\textless \text{PST}\textgreater} as input to the subsequent layer.
Prepending the {\textless \text{SET}\textgreater} aims to refine the sentence embedding so that subsequent tokens can better capture the sentence's semantics.
This procedure is formalized as follows:
\begin{equation}
    \textbf{h}^{l} = {\rm LLM}^{l-1}(f(\textbf{h}^{l-1})), \quad l\in[2,k] \nonumber
\end{equation}
\vspace{-3em}

\begin{equation}
    \textbf{h}^{l-1} = [h_1^{l-1}, \cdots, h_{i^*}^{l-1}, \cdots, h_n^{l-1}] , l\in[2,k] \nonumber
\end{equation}

\vspace{-2em}
\begin{equation}
    f(\textbf{h}^{l-1}) = [h_1^{l-1}, \cdots, h_n^{l-1}, \cdots, h_n^{l-1}] , l\in[2,k] \nonumber
\end{equation}
where $f(\textbf{h})$ represents the function that operates on $\textbf{h}$. $k \in [2, L]$ denotes the ending layer for the intermediate token prepending and \(i^*\) is the position index of the {\textless \text{PST}\textgreater} token.

\begin{table*}[th]
\centering
\normalsize
\setlength{\tabcolsep}{5pt}
\resizebox{\linewidth}{!}{%
\begin{tabular}{lcccccccccc}
\toprule
\textbf{Method} & \textbf{Params} & \textbf{STS12} & \textbf{STS13} & \textbf{STS14} & \textbf{STS15} & \textbf{STS16} & \textbf{STS-B} & \textbf{SICK-R} & \textbf{Avg.} & \textbf{Time}\\
\midrule

BERT avg & 110M & 30.87 & 59.89 & 47.73 & 60.29 & 63.73 & 47.29 & 58.22 & 52.57 & -\\
BERT prompt & 110M & 60.96  & 73.83 & 62.18 & 71.54 & 68.68 & 70.60 & 67.16 & 67.85 & -\\
ST5-Enc avg & 4.8B & 34.97 & 60.19 & 47.59 & 66.40 & 70.62 & 62.83 & 63.57 & 58.02 & -\\
LLaMA2 avg & 7B & 35.49 & 53.15 & 40.12 & 55.35 & 53.26 & 42.10 &      49.96 & 47.06 & $1.00\times$ \\
LLaMA2 echo & 7B & 52.40 &72.40&61.24&72.67&73.51&65.73& 64.39& 66.05 & $1.67\times$ \\
\midrule
{PromptEOL} & 7B & 58.81 & 77.01 & 66.34 & 73.22 & 73.56 & 71.66 & 69.64 & 70.03 & $1.00\times$ \\
{PromptEOL} + TP (\textbf{\textit{Ours}}) & 7B & \cellcolor{gray!20}66.90 \up{8.09}& \cellcolor{gray!20}83.12 \up{6.11}& \cellcolor{gray!20}74.31 \up{7.97}& \cellcolor{gray!20}79.87 \up{6.65}& \cellcolor{gray!20}80.03 \up{6.47}& \cellcolor{gray!20}80.67 \up{9.01}& \cellcolor{gray!20}75.40 \up{5.76}& \cellcolor{gray!20}77.19 \up{7.16} & $1.04\times$ \\
\midrule
{MetaEOL}  & 7B& 64.16 & 81.61 & 73.09 & 81.11 & 78.94 & 77.96 & 74.86 & 75.96 & $8.17\times$ \\
MetaEOL + TP (\textbf{\textit{Ours}}) & 7B&  \cellcolor{gray!20}66.15 \up{1.99}& \cellcolor{gray!20}82.37 \up{0.76}& \cellcolor{gray!20}74.89 \up{1.80}& \cellcolor{gray!20}83.77 \up{2.66}& \cellcolor{gray!20}81.49 \up{2.55}& \cellcolor{gray!20}81.46 \up{3.50}& \cellcolor{gray!20}75.27 \up{0.41}& \cellcolor{gray!20}77.91 \up{1.95} & $8.29\times$ \\
\midrule
Pretended CoT & 7B& 67.45 & 83.89 & 74.14 & 79.47 & 80.76 & 78.95 & 73.33 & 76.86 & $1.18\times$ \\
Pretended CoT + TP (\textbf{\textit{Ours}})& 7B& \cellcolor{gray!20}68.52 \up{1.07}& \cellcolor{gray!20}83.44 \down{0.45}& \cellcolor{gray!20}75.23 \up{1.09}& \cellcolor{gray!20}79.36 \down{0.11}& \cellcolor{gray!20}81.33 \up{0.57}& \cellcolor{gray!20}80.37 \up{1.42}& \cellcolor{gray!20}74.51 \up{1.18}& \cellcolor{gray!20}77.54 \up{0.68} & $1.20\times$ \\

\midrule
Knowledge  & 7B& 65.60 & 82.82 & 74.48 & 80.75 & 80.13 & 80.34 & 75.89 & 77.14 & $1.17\times$\\
Knowledge + TP (\textbf{\textit{Ours}})& 7B& \cellcolor{gray!20}66.03 \up{0.43}& \cellcolor{gray!20}83.43 \up{0.61}& \cellcolor{gray!20}74.50 \up{0.02}& \cellcolor{gray!20}80.94 \up{0.19}& \cellcolor{gray!20}81.28 \up{1.15}& \cellcolor{gray!20}80.45 \up{0.11}& \cellcolor{gray!20}76.13 \up{0.24}& \cellcolor{gray!20}77.54 \up{0.40} & $1.20\times$ \\
\bottomrule
\end{tabular}
}
\caption{Results on STS tasks (Spearman correlation scaled by 100x) using LLaMA2-7B as backbone. The Time column refers to the ratio of inference time for various prompt methods relative to PromptEOL on the STS-B test dataset, ensuring the same output layer.}
\label{tab:results}
\end{table*}

\subsubsection{Layer Scope for Token Prepending}
After passing through the prepending-enhanced layers, all tokens in the sentence are contextualized and can perceive the complete semantic meaning of the sentence.
Therefore, we do not use intermediate token prepending in the later layers and directly feed the hidden states into the standard Transformer layers of LLMs to obtain the sentence embedding.
Specifically,
\begin{equation}
    \textbf{h}^{l+1} = {\rm LLM}^l(\textbf{h}^{l}), l\in[k, M] \nonumber
\end{equation}
where $M$ is the exit layer, which can be either an intermediate layer or the last layer of the LLM.

\subsection{Early-Exit from Intermediate Layers}
Recent studies~\cite{liu/fantastic,jin/exploring} demonstrate that each layer of LLMs plays a different role, and the embeddings from the last layer are primarily used for prediction and contain weaker semantic information.
Thus, we propose the early-exit strategy, which uses embeddings from intermediate layers instead of the last layer to serve as sentence embeddings.
We use the validation set to determine which layer of embedding to use, and the overhead of this process is light.
Another advantage of the early-exit strategy is that we can obtain sentence embeddings more quickly during the testing phase.

\section{Experiments}

\subsection{Datasets and Experimental Settings}
We evaluate sentence embeddings on seven semantic textual similarity (STS) datasets, including STS 2012-2016~\cite{DBLP:conf/semeval/AgirreCDG12,DBLP:conf/starsem/AgirreCDGG13,DBLP:conf/semeval/AgirreBCCDGGMRW14,DBLP:conf/semeval/AgirreBCCDGGLMM15,DBLP:conf/semeval/AgirreBCDGMRW16}, STS-B~\cite{Cer17}, and SICK-R~\cite{DBLP:conf/lrec/MarelliMBBBZ14}.
Each sentence pair in the STS datasets is annotated with a pairwise semantic similarity score from 0 to 5.
We use Spearman correlation as evaluation metric, which measures the rank correlation between the predicted similarity scores and annotated similarity scores using a monotonic function.
We use cosine similarity to compute the predicted similarity scores.

Unless otherwise specified, we use the STS-B development set to determine hyperparameters for TP across all prompt and backbone configurations. In all prompts, the placeholder token {\textless \text{PST}\textgreater} is placed before ``[Text]'' in the template.
We use the output from the 27-th layer for PromptEOL, MetaEOL, and Pretended CoT, and from the penultimate layer for Knowledge Enhancement.
After the 8-th layer, we do not perform token prepending.

\subsection{Baselines}
We combine our method with some baselines to demonstrate effectiveness.
\textbf{BERT avg}~\cite{DBLP:conf/naacl/DevlinCLT19}, \textbf{ST5-Enc avg}~\cite{ni2022sentence}, and \textbf{LLaMA2 avg}~\cite{touvron2023llama} average token embeddings to obtain sentence embeddings using different backbones.
\textbf{LLaMA2 echo}~\cite{springer2024repetition} utlizes the strategy of repetition to obtain sentence embeddings.
\textbf{BERT prompt}~\cite{DBLP:conf/emnlp/JiangJHZWZWHDZ22} proposes a simple and effective prompt to extract sentence embeddings from BERT.
\textbf{PromptEOL}~\cite{jiang2023scaling} first proposes a simple and effective prompt to extract sentence embeddings from LLMs.
\textbf{MetaEOL}~\cite{lei2024meta} leverages a diverse set of meta-task prompts to capture multiple representations of sentences from distinct perspectives.
\textbf{Pretended CoT}~\cite{zhang2024simple} uses CoT to inspire the model to extract sentence embeddings.
\textbf{Knowledge}~\cite{zhang2024simple} explicitly infuses the model with human insights into text summarization.

\begin{table*}[th]
\centering
\normalsize
\setlength{\tabcolsep}{5pt}
\resizebox{\linewidth}{!}{%
\begin{tabular}{lrccccccccc}
\toprule
\textbf{Method} & \textbf{Backbone} & \textbf{STS12} & \textbf{STS13} & \textbf{STS14} & \textbf{STS15} & \textbf{STS16} & \textbf{STS-B} & \textbf{SICK-R} & \textbf{Avg.}\\
\midrule
\midrule
Pretended CoT & LLaMA2-7B &67.45 & 83.89 & 74.14 & 79.47 & 80.76 & 78.95 & 73.33 & 76.86  \\
Pretended CoT + TP (\textbf{\textit{Ours}}) & LLaMA2-7B & \cellcolor{gray!20}68.52 \up{1.07}& \cellcolor{gray!20}83.44 \down{0.45}& \cellcolor{gray!20}75.23 \up{1.09}& \cellcolor{gray!20}79.36 \down{0.11}& \cellcolor{gray!20}81.33 \up{0.57}& \cellcolor{gray!20}80.37 \up{1.42}& \cellcolor{gray!20}74.51 \up{1.18}& \cellcolor{gray!20}77.54 \up{0.68}\\
\midrule
Pretended CoT & LLaMA2-13B & 64.27 & 78.61&69.93&76.37&79.28&75.88&69.04&73.34 \\
Pretended CoT + TP (\textbf{\textit{Ours}}) & LLaMA2-13B & \cellcolor{gray!20}65.65 \up{1.38}& \cellcolor{gray!20}79.50 \up{0.89}& \cellcolor{gray!20}71.01 \up{1.08}& \cellcolor{gray!20}77.27 \up{0.90}& \cellcolor{gray!20}80.07 \up{0.79}& \cellcolor{gray!20}77.36 \up{1.48}& \cellcolor{gray!20}71.51 \up{2.47}& \cellcolor{gray!20}74.62 \up{1.28}\\
\midrule
Pretended CoT & LLaMA3-8B & 66.65&82.60&72.40&79.36&80.86&77.09&73.66&76.09  \\
Pretended CoT + TP (\textbf{\textit{Ours}}) & LLaMA3-8B & \cellcolor{gray!20}66.94 \up{0.29}& \cellcolor{gray!20}83.20 \up{0.60}& \cellcolor{gray!20}73.33 \up{0.93}& \cellcolor{gray!20}79.81 \up{0.45}& \cellcolor{gray!20}81.72 \up{0.86}& \cellcolor{gray!20}78.46 \up{1.37}& \cellcolor{gray!20}73.99 \up{0.33}& \cellcolor{gray!20}76.78 \up{0.69}\\
\midrule
Pretended CoT & Qwen2-7B & 61.64&78.24&70.14&74.44&76.63&76.22&73.30&72.94 \\
Pretended CoT + TP (\textbf{\textit{Ours}}) & Qwen2-7B & \cellcolor{gray!20}65.02 \up{3.38}& \cellcolor{gray!20}79.50 \up{1.26}& \cellcolor{gray!20}71.64 \up{1.50}& \cellcolor{gray!20}77.94 \up{3.5}& \cellcolor{gray!20}79.15 \up{2.52}& \cellcolor{gray!20}78.47 \up{2.25}& \cellcolor{gray!20}74.05 \up{0.75}& \cellcolor{gray!20}75.11 \up{2.17}\\
\midrule
Pretended CoT & Gemma2-9B & 69.50&82.71&74.18&79.64&80.60&78.89&73.60&77.02 \\
Pretended CoT + TP (\textbf{\textit{Ours}}) & Gemma2-9B & \cellcolor{gray!20}69.48\down{0.02}& \cellcolor{gray!20}83.39\up{0.68}& \cellcolor{gray!20}74.32\up{0.14}& \cellcolor{gray!20}80.71\up{1.07}& \cellcolor{gray!20}81.24\up{0.64}& \cellcolor{gray!20}79.24\up{0.35}& \cellcolor{gray!20}74.26\up{0.66}& \cellcolor{gray!20}77.52\up{0.50}\\
\bottomrule
\end{tabular}
}
\caption{Results on STS tasks (Spearman correlation scaled by 100x) using different backbones. Since MetaEOL uses multiple prompts, we use the simple and effective Pretended CoT for our experiments.}
\label{tab:backbone}
\end{table*}

\begin{table*}[th]
    \centering
    \small
    \setlength{\tabcolsep}{25pt}
    \begin{tabular}{l c}
        \toprule
        Prompt Template & STS Avg. \\
        \midrule
        This sentence : {\textless \text{PST}\textgreater} ``[Text]'' means in one word: `` & 77.19 \\
        {\textless \text{PST}\textgreater} This sentence : ``[Text]'' means in one word: `` & 76.35 \\
        This sentence : ``{\textless \text{PST}\textgreater} [Text]'' means in one word: `` & 76.71 \\
        This sentence : `` [Text]'' {\textless \text{PST}\textgreater} means in one word: `` & 75.54 \\
        \midrule
        After thinking step by step , this sentence : {\textless \text{PST}\textgreater} ``[Text]'' means in one word: `` & 77.54  \\
        After thinking step by step , {\textless \text{PST}\textgreater} this sentence :  ``[Text]'' means in one word: `` & 77.81  \\
        After thinking step by step , this sentence : ``{\textless \text{PST}\textgreater} [Text]'' means in one word: `` & 77.51  \\
        After thinking step by step , this sentence : `` [Text]'' {\textless \text{PST}\textgreater} means in one word: `` & 77.44  \\
        \bottomrule
    \end{tabular}
    \caption{Influence of the {\textless \text{PST}\textgreater} token's position in the sentence. We use PromptEOL and Pretended CoT as the prompt.}
    \vspace{-0.5em}
    \label{tab:token_position}
\end{table*}

\subsection{Main Results}

The results of our method on the STS tasks are presented in Table \ref{tab:results}.
Our method consistently outperforms all baselines and non-prompt-based methods perform worse than prompt-based ones.
Among all prompt-based methods across all datasets on LLaMA2-7B, our model shows improvement in 26 out of 28 cases.
This shows our method can be seamlessly integrated with various prompt-based methods without training.
Notably, our method achieves the most significant improvement with PromptEOL, enhancing performance by 7.16. 

The significant improvement in PromptEOL may be because the other three baselines incorporate prior knowledge to understand sentences, whereas PromptEOL relies more on modeling backward dependency to grasp semantics.
Moreover, our method effectively narrows the performance gap between different prompts, improving the model's robustness to prompts.

Another advantage of TP technology is that it introduces minimal additional inference time compare to prompt-based methods. 
We evaluate the inference time by running the LLaMA2-7B model on the STS-Benchmark test dataset, fixing the batch size to 1. 
To mitigate the impact of repeatedly loading the prompt prefix, we employ KV cache.
The comparison results are shown in the Time column of Table \ref{tab:results}. 
We observe that Pretended CoT, Knowledge Enhancement, and MetaEOL incur 1.18, 1.17 and 8.17 times the inference time of PromptEOL, respectively. 
In contrast, the inference time of prompt-based methods with TP technology is within 1.04 times of the original, adding negligible overhead.

\subsection{Evaluation of Different Backbones}

Table \ref{tab:backbone} highlights the performance across various model backbones. In addition to the 7B and 13B versions of LLaMA2 , we evaluate our method on several state-of-the-art decoder-only large language models, including Qwen2-7B \cite{yang2024qwen2}, LLaMA3-8B \cite{dubey2024llama}, and Gemma2-9B \cite{team2024gemma}, using Pretended CoT as the prompt template.

The results demonstrate that our method adapts effectively to a range of large language models, delivering performance gains across different backbones.
Notably, on the Qwen2-7B model, our model achieves an improvement of 2.17 points.
In addition, LLaMA2-13B and LLaMA3-8B do not achieve better performance than LLaMA2-7B.

\begin{figure}[t]
    \centering
    \includegraphics[width=0.4\textwidth]{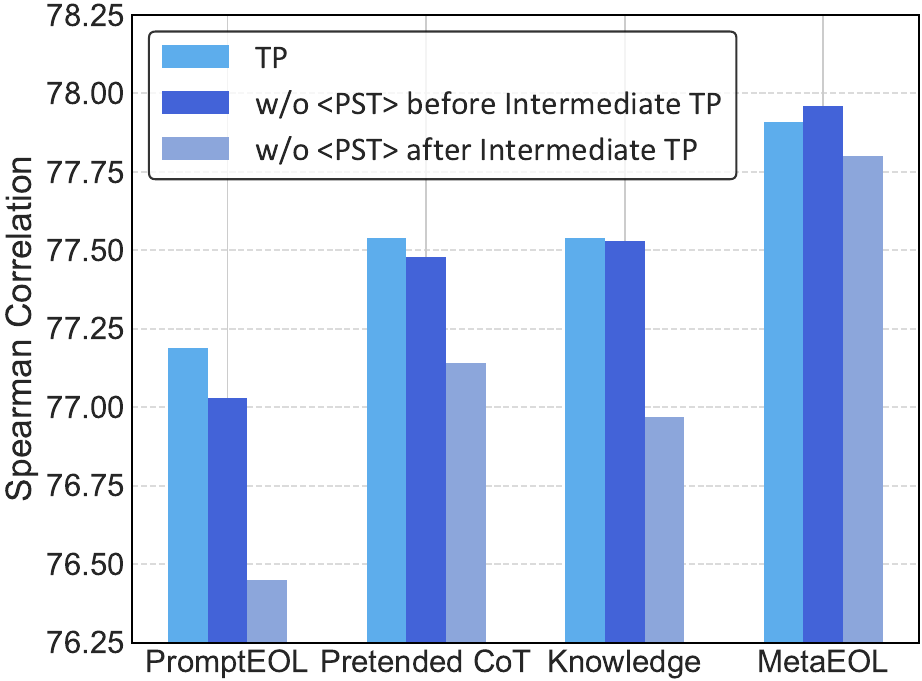}
    \caption{Ablation of {\textless \text{PST}\textgreater} token before and after intermediate token prepending.}
    \label{fig:cls_bar}
\end{figure}

\subsection{Analysis of {\textless \text{PST}\textgreater} Token}

In this section, we analyze the prepended {\textless \text{PST}\textgreater} token in detail using LLaMA2-7B.

\noindent\textbf{Effects of the position of {\textless \text{PST}\textgreater}}\indent We employ PromptEOL and Pretended CoT as the prompt template to examine how the placement of the {\textless \text{PST}\textgreater} token at different positions in the sentence affects performance on STS tasks, as shown in Table \ref{tab:token_position}.
The performance is worst when {\textless \text{PST}\textgreater} token is inserted right after the input text.
When {\textless \text{PST}\textgreater} token is placed before the text, performance fluctuation is small.
The optimal position of {\textless \text{PST}\textgreater} token varies depending on the prompt, typically positioning it close to the text.
To avoid the additional overhead of searching the position, we simply place {\textless \text{PST}\textgreater} token after the colon for all prompts.

\begin{figure*}[t]
    \centering
    \includegraphics[width=1.0\textwidth]{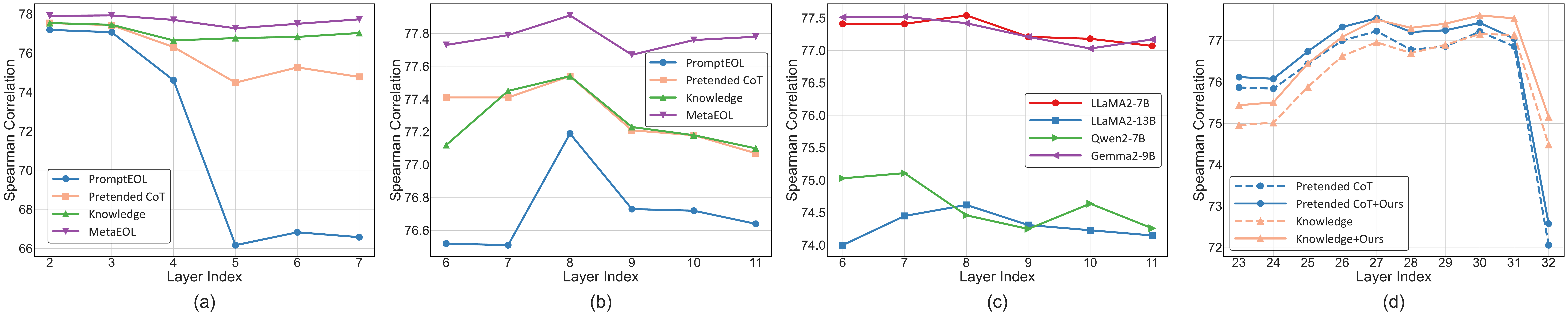}
    \caption{Effects of layer scope for intermediate token prepending and early-exit layer. The reported Spearman correlation is the average across the seven STS tasks. (a) The influence of start layer for the intermediate token prepending. (b) The influence of end layer $k$ for the intermediate token prepending. (c) The token prepending ending layer $k$ on different backbones. (d) The influence of exit layer $M$ for sentence embeddings.}
    \label{fig:layer_analysis}
\end{figure*}

\noindent\textbf{Effectiveness of retaining the {\textless \text{PST}\textgreater} token before and after intermediate TP} 
We ablate the {\textless \text{PST}\textgreater} token before and after intermediate token prepending to show the effectiveness of retaining the {\textless \text{PST}\textgreater} token.
Ablating the {\textless \text{PST}\textgreater} token before the intermediate token prepending is equivalent to removing the initial token prepending and directly performing intermediate token prepending.

The results as shown in Figure \ref{fig:cls_bar}.
Without {\textless \text{PST}\textgreater} before intermediate token prepending generally results in  a slight decrease in performance across most prompts.
This is because the initial {\textless \text{PST}\textgreater} token does not carry semantic information, and its main purpose is to maintain the same input sequence length across all layers.
In contrast, ablating the {\textless \text{PST}\textgreater} token after intermediate token prepending has a more pronounced negative impact.
This may be because the representations are aligned with this input pattern, and modifying the input could lead to a decrease in performance.

\noindent\textbf{Influence of {\textless \text{PST}\textgreater} token initialization} \indent We use Pretended CoT to investigate the effect of various initialization methods for the {\textless \text{PST}\textgreater} token parameters in the embedding layer. We evaluate five initialization techniques: all 0, all 1, uniform distribution within the range [0,1], Gaussian distribution, and using existing token parameters. For the existing token initialization, we select the embedding of the space character, allowing the model to interpret the {\textless \text{PST}\textgreater} token as a space, thereby minimizing its impact on the whole sentence's meaning.

As shown in Table \ref{tab:initialization}, the variation among different initialization is minimal, with the maximum performance difference in STS tasks being just 0.01. This suggests that our method remains robust regardless of the {\textless \text{PST}\textgreater} token's initialization.

\begin{table}[t] \small
    \centering
    \setlength{\tabcolsep}{22pt}
    \begin{tabular}{l c}
        \toprule
        \textbf{Initialization Method} & \textbf{STS Avg.} \\
        \midrule
        All 0 & 77.54  \\
        All 1 & 77.54  \\
        Uniform & 77.53  \\
        Gaussian & 77.54  \\
        Existing token & 77.55 \\
        \bottomrule
    \end{tabular}
    \caption{Influence of the {\textless \text{PST}\textgreater} token's initialization method.}
    \label{tab:initialization}
    \vspace{-0.8em}
\end{table}

\begin{table*}[t]
\centering
\normalsize
\setlength{\tabcolsep}{5pt}
\resizebox{\linewidth}{!}{%
\begin{tabular}{lcccccccccc}
\toprule
\textbf{Method} & \textbf{MR} & \textbf{CR} & \textbf{SUBJ} & \textbf{MPQA} & \textbf{SST2} & \textbf{TREC} & \textbf{MRPC} & \textbf{Avg.}\\
\midrule
\midrule
PromptEOL & 90.63 &92.87 &96.32 &91.19 &95.00 &95.40 &75.19 &90.94  \\
PromptEOL + TP (\textbf{\textit{Ours}}) &  \cellcolor{gray!20}90.90\up{0.27}& \cellcolor{gray!20}93.35 \up{0.48}& \cellcolor{gray!20}96.58 \up{0.26}& \cellcolor{gray!20}91.51 \up{0.32}& \cellcolor{gray!20}95.50 \up{0.50}& \cellcolor{gray!20}96.00 \up{0.60}& \cellcolor{gray!20}76.12 \up{0.93}& \cellcolor{gray!20}91.42 \up{0.48}\\
\midrule
Pretended CoT & 90.10&92.24&96.32&91.54&95.11&94.20&75.77&90.75  \\
Pretended CoT + TP (\textbf{\textit{Ours}}) &  \cellcolor{gray!20}90.45\up{0.35}& \cellcolor{gray!20}92.61\up{0.37}& \cellcolor{gray!20}96.52\up{0.20}& \cellcolor{gray!20}91.59\up{0.05}& \cellcolor{gray!20}95.77 \up{0.66}& \cellcolor{gray!20}96.00\up{1.80}& \cellcolor{gray!20}76.81\up{1.04}& \cellcolor{gray!20}91.39\up{0.64}\\
\midrule
Knowledge & 89.84&93.03&96.21&91.54&94.78&97.20&73.91&90.93   \\
Knowledge + TP (\textbf{\textit{Ours}}) &  \cellcolor{gray!20}90.39\up{0.55}& \cellcolor{gray!20}93.32\up{0.29}& \cellcolor{gray!20}96.31\up{0.10}& \cellcolor{gray!20}91.56\up{0.02}& \cellcolor{gray!20}94.51\down{0.27}& \cellcolor{gray!20}97.60\up{0.40}& \cellcolor{gray!20}76.06\up{2.15}& \cellcolor{gray!20}91.39\up{0.46}\\

\bottomrule
\end{tabular}
}
\caption{Results (accuracy scaled by 100x) on transfer learning tasks using LLaMA2-7B.}
\vspace{-0.5em}
\label{tab:transfer_tasks}

\end{table*}

\subsection{Analysis of Layer Scope for TP}
In this section, we analyze the effects of layer scope for token prepending in detail.

\noindent\textbf{Influence of start and end layer for TP} \indent In Figure \ref{fig:layer_analysis}(a) and (b), we explore the impact of the start and end layer for intermediate token prepending on LLaMA2-7B.

As illustrated in Figure \ref{fig:layer_analysis}(a), performance is suboptimal if intermediate token prepending does not begin at the second layer.
This is because the {\textless \text{PST}\textgreater} token is randomly initialized and lacks semantic information.
Thus, we need to replace it with semantically meaningful tokens at the early layers of the LLMs to mitigate this issue.

Furthermore, our results indicate that halting token prepending after the 8-th layer yields the best performance across all prompts used.

\noindent\textbf{Influence of end layer for intermediate TP on different backbones} We further examine the optimal layer to terminate token prepending across different backbones.
As shown in Figure \ref{fig:layer_analysis}(c), for the LLaMA2-7B and LLaMA2-13B models, stopping token prepending at the 8-th yields the best performance. While for the Qwen2-7B and Gemma2-9B models, the optimal ending layer is 7-th. This suggests that, for most decoder-only LLMs, modeling shallow-layer backward dependencies is crucial for enhancing sentence comprehension.
The best layer for ending token prepending is similar across different backbones, typically falling within the 7-th to 8-th layer range.

\subsection{Influence of Exit Layers}
We examine the impact of exit layers in LLaMA2-7B using Pretended CoT and Knowledge Enhancement. 
As illustrated in Figure \ref{fig:layer_analysis}(d), our model consistently improves both Pretended CoT and Knowledge Enhancement across all layers and configurations.
Furthermore, Pretended CoT and Knowledge Enhancement exhibit greater variability in performance across different layers compared to ours.
This implies that the our method offers a more stable representational quality across layers.

Notably, employing the output of the model's last layer is consistently suboptimal for STS tasks, consistent with prior research~\cite{li2024bellm, lei2024meta}.
Pretended CoT achieves optimal performance at the sixth-to-last layer, while Knowledge Enhancement peaks at the second-to-last layer.
This variation suggests that the optimal layer can shift depending on the prompt.

\subsection{Transfer Learning Tasks} \indent We further evaluate the performance of our model on transfer learning tasks. 
We use standard transfer learning tasks provided by SentEval, including MR \cite{pang2005seeing}, CR \cite{hu2004mining}, SUBJ \cite{pang2004sentimental}, MPQA \cite{wiebe2005annotating}, SST-2 \cite{socher2013recursive}, TREC \cite{voorhees2000building}, and MRPC \cite{dolan2005automatically}.
For each task, we use the sentence embeddings generated by the model as features to train logistic regression classifiers. 

The results for the transfer tasks, shown in Table~\ref{tab:transfer_tasks}, demonstrate that our method consistently outperforms all baselines, with improvements in 20 out of 21 cases across all datasets. 
This indicates that TP cultivates generalized sentence embeddings that perform outstandingly across various tasks.
Pretended CoT and Knowledge Enhancement do not surpass the performance of PromptEOL, indicating that they are not consistently effective in enhancing performance across tasks.

Additionally, we find that ending token prepending at deeper layers typically between layer index 14 and 21, enhances performance on transfer tasks.
This phenomenon differs significantly from the optimal layer for STS tasks, suggesting that transfer tasks benefit from additional layers to effectively model backward dependencies.

\begin{figure}[t]
    \centering
    \includegraphics[width=0.4\textwidth]{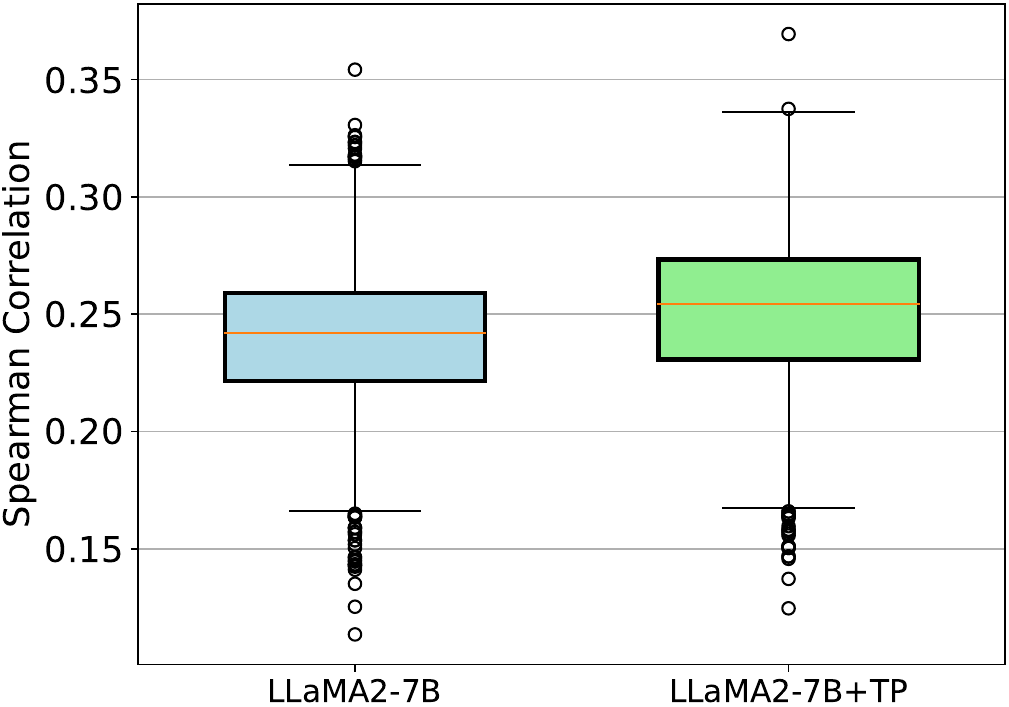}
    \caption{Box plot of the sentence-level Spearman correlation on the STS-B test set using Pretended CoT prompt.}
    \label{fig:box_spearman}
    \vspace{-0.8em}
\end{figure}

\subsection{Evaluation of Capturing Dependencies in Contexts}
We quantitatively analyse whether our proposed method enhances the ability of LLMs to capture dependencies in contexts on the STS-B test set using LLaMA2-7B.
For both models, we follow \citet{ethayarajh2019contextual} by selecting the last token as the pivot token.
We then compute the Spearman correlation between the pivot token and the remaining tokens in each sentence to assess their dependency-capturing capabilities.
The results are shown in a box plot in Figure \ref{fig:box_spearman}.

The average sentence-level Spearman score for LLaMA2-7B and LLaMA2-7B+TP are 23.97 and 25.11, respectively.
The results indicate that our method achieves an improved capability to capture backward dependencies compared to vanilla LLaMA2-7B model.
This suggests that token prepending offers benefits for enhancing the ability of LLMs to capture dependencies in contexts.

\section{Conclusion}
\vspace{-0.4em}
In this paper, we introduce Token Prepending technique, a plug-and-play approach for deriving high-quality sentence embeddings from autogressive LLMs without requiring any training and data. 
By intervening in the inputs to Transformer layers, TP enhances the ability of autoregressive LLMs to capture backward dependencies. 
Moreover, TP involves simply prepending a single token to the sentence, which adds negligible inference cost and can seamlessly integrate with prompt-based methods. 
Our extensive experiments demonstrate that TP technique can effectively and generally elicit sentence embeddings across a range of LLMs with varying architectures and parameter sizes, achieving outstanding performance on both STS datasets and transfer learning tasks.
We find that starting TP from the first layer yields optimal results, and the best stopping point is typically around the 7th or 8th layer for LLMs with about 7B parameters.

\section*{Limitations}

Although Token Prepending is a training-free technique, it requires tuning two hyperparameters (i.e., end layer for intermediate token prepending and exit layer) to achieve optimal sentence embeddings. 
Our results show that the best hyperparameters for TP vary based on the model, dataset, and prompt, which may increase adaptation costs when applying it to new scenarios.

\section*{Acknowledgments}

We would like to thank the anonymous reviewers for their insightful comments. 
This work is supported by the National Natural Science Foundation of China under Grants Nos. 62441225, 61972192, 62172208, 61906085. 
This work is partially supported by Collaborative Innovation Center of Novel Software Technology and Industrialization.
This work is supported by the Fundamental Research Funds for the Central Universities under Grant No.14380001.

\bibliography{acl_latex.bib}

\begin{thebibliography}{44}
\providecommand{\natexlab}[1]{#1}

\bibitem[{Agirre et~al.(2015)Agirre, Banea, Cardie, Cer, Diab,
  Gonzalez{-}Agirre, Guo, Lopez{-}Gazpio, Maritxalar, Mihalcea, Rigau, Uria,
  and Wiebe}]{DBLP:conf/semeval/AgirreBCCDGGLMM15}
Eneko Agirre, Carmen Banea, Claire Cardie, Daniel~M. Cer, Mona~T. Diab, Aitor
  Gonzalez{-}Agirre, Weiwei Guo, I{\~{n}}igo Lopez{-}Gazpio, Montse Maritxalar,
  Rada Mihalcea, German Rigau, Larraitz Uria, and Janyce Wiebe. 2015.
\newblock \href {https://doi.org/10.18653/V1/S15-2045} {Semeval-2015 task 2:
  Semantic textual similarity, english, spanish and pilot on interpretability}.
\newblock In \emph{Proceedings of the 9th International Workshop on Semantic
  Evaluation, SemEval@NAACL-HLT 2015}, pages 252--263. The Association for
  Computer Linguistics.

\bibitem[{Agirre et~al.(2014)Agirre, Banea, Cardie, Cer, Diab,
  Gonzalez{-}Agirre, Guo, Mihalcea, Rigau, and
  Wiebe}]{DBLP:conf/semeval/AgirreBCCDGGMRW14}
Eneko Agirre, Carmen Banea, Claire Cardie, Daniel~M. Cer, Mona~T. Diab, Aitor
  Gonzalez{-}Agirre, Weiwei Guo, Rada Mihalcea, German Rigau, and Janyce Wiebe.
  2014.
\newblock \href {https://doi.org/10.3115/V1/S14-2010} {Semeval-2014 task 10:
  Multilingual semantic textual similarity}.
\newblock In \emph{Proceedings of the 8th International Workshop on Semantic
  Evaluation, SemEval@COLING 2014}, pages 81--91. The Association for Computer
  Linguistics.

\bibitem[{Agirre et~al.(2016)Agirre, Banea, Cer, Diab, Gonzalez{-}Agirre,
  Mihalcea, Rigau, and Wiebe}]{DBLP:conf/semeval/AgirreBCDGMRW16}
Eneko Agirre, Carmen Banea, Daniel~M. Cer, Mona~T. Diab, Aitor
  Gonzalez{-}Agirre, Rada Mihalcea, German Rigau, and Janyce Wiebe. 2016.
\newblock \href {https://doi.org/10.18653/V1/S16-1081} {Semeval-2016 task 1:
  Semantic textual similarity, monolingual and cross-lingual evaluation}.
\newblock In \emph{Proceedings of the 10th International Workshop on Semantic
  Evaluation, SemEval@NAACL-HLT 2016}, pages 497--511. The Association for
  Computer Linguistics.

\bibitem[{Agirre et~al.(2012)Agirre, Cer, Diab, and
  Gonzalez{-}Agirre}]{DBLP:conf/semeval/AgirreCDG12}
Eneko Agirre, Daniel~M. Cer, Mona~T. Diab, and Aitor Gonzalez{-}Agirre. 2012.
\newblock \href {https://aclanthology.org/S12-1051/} {Semeval-2012 task 6: {A}
  pilot on semantic textual similarity}.
\newblock In \emph{Proceedings of the 6th International Workshop on Semantic
  Evaluation, SemEval@NAACL-HLT 2012}, pages 385--393. The Association for
  Computer Linguistics.

\bibitem[{Agirre et~al.(2013)Agirre, Cer, Diab, Gonzalez{-}Agirre, and
  Guo}]{DBLP:conf/starsem/AgirreCDGG13}
Eneko Agirre, Daniel~M. Cer, Mona~T. Diab, Aitor Gonzalez{-}Agirre, and Weiwei
  Guo. 2013.
\newblock \href {https://aclanthology.org/S13-1004/} {*sem 2013 shared task:
  Semantic textual similarity}.
\newblock In \emph{Proceedings of the Second Joint Conference on Lexical and
  Computational Semantics, *SEM 2013}, pages 32--43. Association for
  Computational Linguistics.

\bibitem[{BehnamGhader et~al.(2024)BehnamGhader, Adlakha, Mosbach, Bahdanau,
  Chapados, and Reddy}]{behnamghader2024llm2vec}
Parishad BehnamGhader, Vaibhav Adlakha, Marius Mosbach, Dzmitry Bahdanau,
  Nicolas Chapados, and Siva Reddy. 2024.
\newblock Llm2vec: Large language models are secretly powerful text encoders.
\newblock \emph{arXiv preprint arXiv:2404.05961}.

\bibitem[{Brown(2020)}]{brown2020language}
Tom~B Brown. 2020.
\newblock Language models are few-shot learners.
\newblock \emph{arXiv preprint ArXiv:2005.14165}.

\bibitem[{Cer et~al.(2017)Cer, Diab, Agirre, Lopez{-}Gazpio, and
  Specia}]{Cer17}
Daniel~M. Cer, Mona~T. Diab, Eneko Agirre, I{\~{n}}igo Lopez{-}Gazpio, and
  Lucia Specia. 2017.
\newblock \href {https://arxiv.org/abs/1708.00055} {Semeval-2017 task 1:
  Semantic textual similarity - multilingual and cross-lingual focused
  evaluation}.
\newblock \emph{CoRR}, abs/1708.00055.

\bibitem[{Chanchani and Huang(2023)}]{DBLP:conf/acl/ChanchaniH23}
Sachin Chanchani and Ruihong Huang. 2023.
\newblock \href {https://doi.org/10.18653/V1/2023.ACL-LONG.882}
  {Composition-contrastive learning for sentence embeddings}.
\newblock In \emph{Proceedings of the 61st Annual Meeting of the Association
  for Computational Linguistics (Volume 1: Long Papers)}, pages 15836--15848.
  Association for Computational Linguistics.

\bibitem[{Cheng et~al.(2025)Cheng, Wang, Fu, Jiang, Yin, Wang, and
  Gu}]{cheng2025contrastive}
Zifeng Cheng, Zhonghui Wang, Yuchen Fu, Zhiwei Jiang, Yafeng Yin, Cong Wang,
  and Qing Gu. 2025.
\newblock Contrastive prompting enhances sentence embeddings in llms through
  inference-time steering.
\newblock \emph{arXiv preprint arXiv:2505.12831}.

\bibitem[{Devlin et~al.(2019)Devlin, Chang, Lee, and
  Toutanova}]{DBLP:conf/naacl/DevlinCLT19}
Jacob Devlin, Ming{-}Wei Chang, Kenton Lee, and Kristina Toutanova. 2019.
\newblock \href {https://doi.org/10.18653/V1/N19-1423} {{BERT:} pre-training of
  deep bidirectional transformers for language understanding}.
\newblock In \emph{Proceedings of the 2019 Conference of the North American
  Chapter of the Association for Computational Linguistics: Human Language
  Technologies, {NAACL-HLT} 2019}, pages 4171--4186.

\bibitem[{Dolan and Brockett(2005)}]{dolan2005automatically}
Bill Dolan and Chris Brockett. 2005.
\newblock Automatically constructing a corpus of sentential paraphrases.
\newblock In \emph{Third international workshop on paraphrasing (IWP2005)}.

\bibitem[{Dubey et~al.(2024)Dubey, Jauhri, Pandey, Kadian, Al-Dahle, Letman,
  Mathur, Schelten, Yang, Fan et~al.}]{dubey2024llama}
Abhimanyu Dubey, Abhinav Jauhri, Abhinav Pandey, Abhishek Kadian, Ahmad
  Al-Dahle, Aiesha Letman, Akhil Mathur, Alan Schelten, Amy Yang, Angela Fan,
  et~al. 2024.
\newblock The llama 3 herd of models.
\newblock \emph{arXiv preprint arXiv:2407.21783}.

\bibitem[{Ethayarajh(2019)}]{ethayarajh2019contextual}
Kawin Ethayarajh. 2019.
\newblock How contextual are contextualized word representations? comparing the
  geometry of bert, elmo, and gpt-2 embeddings.
\newblock In \emph{Proceedings of the 2019 Conference on Empirical Methods in
  Natural Language Processing and the 9th International Joint Conference on
  Natural Language Processing (EMNLP-IJCNLP)}, pages 55--65.

\bibitem[{Gao et~al.(2021)Gao, Yao, and Chen}]{gao2021simcse}
Tianyu Gao, Xingcheng Yao, and Danqi Chen. 2021.
\newblock Simcse: Simple contrastive learning of sentence embeddings.
\newblock In \emph{Proceedings of the 2021 Conference on Empirical Methods in
  Natural Language Processing}, pages 6894--6910.

\bibitem[{Hu and Liu(2004)}]{hu2004mining}
Minqing Hu and Bing Liu. 2004.
\newblock Mining and summarizing customer reviews.
\newblock In \emph{Proceedings of the tenth ACM SIGKDD international conference
  on Knowledge discovery and data mining}, pages 168--177.

\bibitem[{Jiang et~al.(2023)Jiang, Huang, Luan, Wang, and
  Zhuang}]{jiang2023scaling}
Ting Jiang, Shaohan Huang, Zhongzhi Luan, Deqing Wang, and Fuzhen Zhuang. 2023.
\newblock Scaling sentence embeddings with large language models.
\newblock \emph{arXiv preprint arXiv:2307.16645}.

\bibitem[{Jiang et~al.(2022)Jiang, Jiao, Huang, Zhang, Wang, Zhuang, Wei,
  Huang, Deng, and Zhang}]{DBLP:conf/emnlp/JiangJHZWZWHDZ22}
Ting Jiang, Jian Jiao, Shaohan Huang, Zihan Zhang, Deqing Wang, Fuzhen Zhuang,
  Furu Wei, Haizhen Huang, Denvy Deng, and Qi~Zhang. 2022.
\newblock \href {https://doi.org/10.18653/V1/2022.EMNLP-MAIN.603} {Promptbert:
  Improving {BERT} sentence embeddings with prompts}.
\newblock In \emph{Proceedings of the 2022 Conference on Empirical Methods in
  Natural Language Processing, {EMNLP} 2022}, pages 8826--8837.

\bibitem[{Jin et~al.(2024{\natexlab{a}})Jin, Yu, Huang, Zeng, Wang, Hua, Zhao,
  Mei, Meng, Ding, Yang, Du, and Zhang}]{jin/exploring}
Mingyu Jin, Qinkai Yu, Jingyuan Huang, Qingcheng Zeng, Zhenting Wang, Wenyue
  Hua, Haiyan Zhao, Kai Mei, Yanda Meng, Kaize Ding, Fan Yang, Mengnan Du, and
  Yongfeng Zhang. 2024{\natexlab{a}}.
\newblock \href {https://doi.org/10.48550/ARXIV.2404.07066} {Exploring concept
  depth: How large language models acquire knowledge at different layers?}
\newblock \emph{CoRR}, abs/2404.07066.

\bibitem[{Jin et~al.(2024{\natexlab{b}})Jin, Yu, Huang, Zeng, Wang, Hua, Zhao,
  Mei, Meng, Ding et~al.}]{jin2024exploring}
Mingyu Jin, Qinkai Yu, Jingyuan Huang, Qingcheng Zeng, Zhenting Wang, Wenyue
  Hua, Haiyan Zhao, Kai Mei, Yanda Meng, Kaize Ding, et~al. 2024{\natexlab{b}}.
\newblock Exploring concept depth: How large language models acquire knowledge
  and concept at different layers?
\newblock \emph{arXiv preprint arXiv:2404.07066}.

\bibitem[{Lee et~al.(2024)Lee, Roy, Xu, Raiman, Shoeybi, Catanzaro, and
  Ping}]{NV-Embed/Lee}
Chankyu Lee, Rajarshi Roy, Mengyao Xu, Jonathan Raiman, Mohammad Shoeybi, Bryan
  Catanzaro, and Wei Ping. 2024.
\newblock \href {https://doi.org/10.48550/ARXIV.2405.17428} {Nv-embed: Improved
  techniques for training llms as generalist embedding models}.
\newblock \emph{CoRR}, abs/2405.17428.

\bibitem[{Lei et~al.(2024)Lei, Wu, Zhou, Shen, Cao, Tao, and
  Yates}]{lei2024meta}
Yibin Lei, Di~Wu, Tianyi Zhou, Tao Shen, Yu~Cao, Chongyang Tao, and Andrew
  Yates. 2024.
\newblock Meta-task prompting elicits embedding from large language models.
\newblock \emph{arXiv preprint arXiv:2402.18458}.

\bibitem[{Li and Li(2023)}]{li2023angle}
Xianming Li and Jing Li. 2023.
\newblock Angle-optimized text embeddings.
\newblock \emph{arXiv preprint arXiv:2309.12871}.

\bibitem[{Li and Li(2024)}]{li2024bellm}
Xianming Li and Jing Li. 2024.
\newblock Bellm: Backward dependency enhanced large language model for sentence
  embeddings.
\newblock In \emph{Proceedings of the 2024 Conference of the North American
  Chapter of the Association for Computational Linguistics: Human Language
  Technologies (Volume 1: Long Papers)}, pages 792--804.

\bibitem[{Liu et~al.(2024{\natexlab{a}})Liu, Trager, Achille, Perera, Zancato,
  and Soatto}]{liumeaning}
Tian~Yu Liu, Matthew Trager, Alessandro Achille, Pramuditha Perera, Luca
  Zancato, and Stefano Soatto. 2024{\natexlab{a}}.
\newblock Meaning representations from trajectories in autoregressive models.
\newblock In \emph{The Twelfth International Conference on Learning
  Representations}.

\bibitem[{Liu et~al.(2024{\natexlab{b}})Liu, Kong, Liu, and
  Sun}]{liu2024fantastic}
Zhu Liu, Cunliang Kong, Ying Liu, and Maosong Sun. 2024{\natexlab{b}}.
\newblock Fantastic semantics and where to find them: Investigating which
  layers of generative llms reflect lexical semantics.
\newblock \emph{arXiv preprint arXiv:2403.01509}.

\bibitem[{Liu et~al.(2024{\natexlab{c}})Liu, Kong, Liu, and
  Sun}]{liu/fantastic}
Zhu Liu, Cunliang Kong, Ying Liu, and Maosong Sun. 2024{\natexlab{c}}.
\newblock \href {https://doi.org/10.48550/ARXIV.2403.01509} {Fantastic
  semantics and where to find them: Investigating which layers of generative
  llms reflect lexical semantics}.
\newblock \emph{CoRR}, abs/2403.01509.

\bibitem[{Marelli et~al.(2014)Marelli, Menini, Baroni, Bentivogli, Bernardi,
  and Zamparelli}]{DBLP:conf/lrec/MarelliMBBBZ14}
Marco Marelli, Stefano Menini, Marco Baroni, Luisa Bentivogli, Raffaella
  Bernardi, and Roberto Zamparelli. 2014.
\newblock \href
  {http://www.lrec-conf.org/proceedings/lrec2014/summaries/363.html} {A {SICK}
  cure for the evaluation of compositional distributional semantic models}.
\newblock In \emph{Proceedings of the Ninth International Conference on
  Language Resources and Evaluation, {LREC} 2014}, pages 216--223.

\bibitem[{Muennighoff et~al.(2024)Muennighoff, Su, Wang, Yang, Wei, Yu, Singh,
  and Kiela}]{generative/muennighoff}
Niklas Muennighoff, Hongjin Su, Liang Wang, Nan Yang, Furu Wei, Tao Yu,
  Amanpreet Singh, and Douwe Kiela. 2024.
\newblock \href {https://doi.org/10.48550/ARXIV.2402.09906} {Generative
  representational instruction tuning}.
\newblock \emph{CoRR}, abs/2402.09906.

\bibitem[{Muennighoff et~al.(2022)Muennighoff, Tazi, Magne, and
  Reimers}]{muennighoff2022mteb}
Niklas Muennighoff, Nouamane Tazi, Lo{\"\i}c Magne, and Nils Reimers. 2022.
\newblock Mteb: Massive text embedding benchmark.
\newblock \emph{arXiv preprint arXiv:2210.07316}.

\bibitem[{Ni et~al.(2022{\natexlab{a}})Ni, Abrego, Constant, Ma, Hall, Cer, and
  Yang}]{ni2022sentence}
Jianmo Ni, Gustavo~Hernandez Abrego, Noah Constant, Ji~Ma, Keith Hall, Daniel
  Cer, and Yinfei Yang. 2022{\natexlab{a}}.
\newblock Sentence-t5: Scalable sentence encoders from pre-trained text-to-text
  models.
\newblock In \emph{Findings of the Association for Computational Linguistics:
  ACL 2022}, pages 1864--1874.

\bibitem[{Ni et~al.(2022{\natexlab{b}})Ni, {\'{A}}brego, Constant, Ma, Hall,
  Cer, and Yang}]{DBLP:conf/acl/NiACMHCY22}
Jianmo Ni, Gustavo~Hern{\'{a}}ndez {\'{A}}brego, Noah Constant, Ji~Ma, Keith~B.
  Hall, Daniel Cer, and Yinfei Yang. 2022{\natexlab{b}}.
\newblock \href {https://doi.org/10.18653/V1/2022.FINDINGS-ACL.146}
  {Sentence-t5: Scalable sentence encoders from pre-trained text-to-text
  models}.
\newblock In \emph{Findings of the Association for Computational Linguistics:
  {ACL} 2022}, pages 1864--1874.

\bibitem[{Pang and Lee(2004)}]{pang2004sentimental}
Bo~Pang and Lillian Lee. 2004.
\newblock A sentimental education: Sentiment analysis using subjectivity
  summarization based on minimum cuts.
\newblock In \emph{Proceedings of the 42nd Annual Meeting of the Association
  for Computational Linguistics (ACL-04)}, pages 271--278.

\bibitem[{Pang and Lee(2005)}]{pang2005seeing}
Bo~Pang and Lillian Lee. 2005.
\newblock Seeing stars: Exploiting class relationships for sentiment
  categorization with respect to rating scales.
\newblock In \emph{Proceedings of the 43rd Annual Meeting of the Association
  for Computational Linguistics (ACL’05)}, pages 115--124.

\bibitem[{Raffel et~al.(2020)Raffel, Shazeer, Roberts, Lee, Narang, Matena,
  Zhou, Li, and Liu}]{DBLP:journals/jmlr/RaffelSRLNMZLL20}
Colin Raffel, Noam Shazeer, Adam Roberts, Katherine Lee, Sharan Narang, Michael
  Matena, Yanqi Zhou, Wei Li, and Peter~J. Liu. 2020.
\newblock \href {http://jmlr.org/papers/v21/20-074.html} {Exploring the limits
  of transfer learning with a unified text-to-text transformer}.
\newblock \emph{J. Mach. Learn. Res.}, 21:140:1--140:67.

\bibitem[{Socher et~al.(2013)Socher, Perelygin, Wu, Chuang, Manning, Ng, and
  Potts}]{socher2013recursive}
Richard Socher, Alex Perelygin, Jean Wu, Jason Chuang, Christopher~D Manning,
  Andrew~Y Ng, and Christopher Potts. 2013.
\newblock Recursive deep models for semantic compositionality over a sentiment
  treebank.
\newblock In \emph{Proceedings of the 2013 conference on empirical methods in
  natural language processing}, pages 1631--1642.

\bibitem[{Springer et~al.(2024)Springer, Kotha, Fried, Neubig, and
  Raghunathan}]{springer2024repetition}
Jacob~Mitchell Springer, Suhas Kotha, Daniel Fried, Graham Neubig, and Aditi
  Raghunathan. 2024.
\newblock Repetition improves language model embeddings.
\newblock \emph{arXiv preprint arXiv:2402.15449}.

\bibitem[{Su et~al.(2023)Su, Shi, Kasai, Wang, Hu, Ostendorf, Yih, Smith,
  Zettlemoyer, and Yu}]{DBLP:conf/acl/SuSKWHOYSZ023}
Hongjin Su, Weijia Shi, Jungo Kasai, Yizhong Wang, Yushi Hu, Mari Ostendorf,
  Wen{-}tau Yih, Noah~A. Smith, Luke Zettlemoyer, and Tao Yu. 2023.
\newblock \href {https://doi.org/10.18653/V1/2023.FINDINGS-ACL.71} {One
  embedder, any task: Instruction-finetuned text embeddings}.
\newblock In \emph{Findings of the Association for Computational Linguistics:
  {ACL} 2023}, pages 1102--1121. Association for Computational Linguistics.

\bibitem[{Team et~al.(2024)Team, Riviere, Pathak, Sessa, Hardin, Bhupatiraju,
  Hussenot, Mesnard, Shahriari, Ram{\'e} et~al.}]{team2024gemma}
Gemma Team, Morgane Riviere, Shreya Pathak, Pier~Giuseppe Sessa, Cassidy
  Hardin, Surya Bhupatiraju, L{\'e}onard Hussenot, Thomas Mesnard, Bobak
  Shahriari, Alexandre Ram{\'e}, et~al. 2024.
\newblock Gemma 2: Improving open language models at a practical size.
\newblock \emph{arXiv preprint arXiv:2408.00118}.

\bibitem[{Touvron et~al.(2023)Touvron, Martin, Stone, Albert, Almahairi,
  Babaei, Bashlykov, Batra, Bhargava, Bhosale et~al.}]{touvron2023llama}
Hugo Touvron, Louis Martin, Kevin Stone, Peter Albert, Amjad Almahairi, Yasmine
  Babaei, Nikolay Bashlykov, Soumya Batra, Prajjwal Bhargava, Shruti Bhosale,
  et~al. 2023.
\newblock Llama 2: Open foundation and fine-tuned chat models.
\newblock \emph{arXiv preprint arXiv:2307.09288}.

\bibitem[{Voorhees and Tice(2000)}]{voorhees2000building}
Ellen~M Voorhees and Dawn~M Tice. 2000.
\newblock Building a question answering test collection.
\newblock In \emph{Proceedings of the 23rd annual international ACM SIGIR
  conference on Research and development in information retrieval}, pages
  200--207.

\bibitem[{Wiebe et~al.(2005)Wiebe, Wilson, and Cardie}]{wiebe2005annotating}
Janyce Wiebe, Theresa Wilson, and Claire Cardie. 2005.
\newblock Annotating expressions of opinions and emotions in language.
\newblock \emph{Language resources and evaluation}, 39:165--210.

\bibitem[{Yang et~al.(2024)Yang, Yang, Hui, Zheng, Yu, Zhou, Li, Li, Liu, Huang
  et~al.}]{yang2024qwen2}
An~Yang, Baosong Yang, Binyuan Hui, Bo~Zheng, Bowen Yu, Chang Zhou, Chengpeng
  Li, Chengyuan Li, Dayiheng Liu, Fei Huang, et~al. 2024.
\newblock Qwen2 technical report.
\newblock \emph{arXiv preprint arXiv:2407.10671}.

\bibitem[{Zhang et~al.(2024)Zhang, Chang, and Li}]{zhang2024simple}
Bowen Zhang, Kehua Chang, and Chunping Li. 2024.
\newblock Simple techniques for enhancing sentence embeddings in generative
  language models.
\newblock \emph{arXiv preprint arXiv:2404.03921}.

\end{thebibliography}

\clearpage
\newpage
\appendix

\section{Appendix}

\subsection{Comparison with Bidirectional Attention}

We explore the performance of removing the causal attention mask. 
To this end, we design two types of bidirectional attention masks: 1) enabling bidirectional attention for the last token, and 2) enabling bidirectional attention for the input sentence. 
To ensure fairness, the starting position of the non-causal attention aligns with the position of the prepended {\textless \text{PST}\textgreater} token. 

Using Pretended CoT as the prompt, the results are presented in Table \ref{tab:causal_masks}. 
Both types of bidirectional attention masks result lead to a substantial decrease in performance. 
This observation is consistent with prior research \cite{behnamghader2024llm2vec, li2024bellm}, which indicates that, due to the inductive bias of autoregressive large language models, employing a bidirectional attention mechanism tends to reduce model performance.

\subsection{Multi-Task Evaluation}

We evaluate the TP technique across 12 classification datasets, 3 pair classification datasets, 4 reranking datasets, 11 clustering datasets, 1 summarization dataset, and 1 additional STS dataset.

\begin{table}[th]
    \centering
    \small
    \setlength{\tabcolsep}{8pt}
    \begin{tabular}{l c}
        \toprule
        \textbf{Initialization Method} & \textbf{STS Avg.} \\
        \midrule
        Vanilla LLM & 76.86  \\
        TP (Ours) & 77.54  \\
        Bidirectional Attention (Last token) & 53.06  \\
        Bidirectional Attention (Input Sentence) & 43.70  \\
        \bottomrule
    \end{tabular}
    \caption{Influence of the modified attention masks.}
    \label{tab:causal_masks}
\end{table}

\begin{table}[th]
    \centering

    \begin{center}
    \setlength{\tabcolsep}{1pt}
    \begin{small}
    \begin{tabular}{c|ccc|cccccccc}
    \hline
    \textbf{Method} & \textbf{PromptEOL} & \textbf{PromptEOL+TP}      \\
    \hline
    
    AmazonCounterfactual      & 70.83                         & 71.71\\
    AmazonPolarity            & 88.48                         & 94.57\\
    AmazonReviews             & 46.03                         & 47.77\\
    Banking77                 & 78.94                         & 82.24\\
    Emotion                   & 48.35                         & 51.05\\
    Imdb                      & 79.10                         & 81.44\\
    MassiveIntent             & 72.49                         & 75.22\\
    MassiveScenario           & 75.41                         & 78.69\\
    MTOPDomain                & 90.49                         & 93.63\\
    MTOPIntent                & 81.48                         & 83.16\\
    ToxicConversations        & 64.51                         & 68.68\\
    TweetSentimentExtraction  & 60.55                         & 61.15\\
     
    \hline
    \bf{Average (12)}         & 71.39                         & 74.11\\
    \hline
    \end{tabular}
    \caption{Results (accuracy scaled by 100x) on classification datasets using LLaMA2-7B.}
    \label{table:mteb_classification}
    \end{small}
    \end{center}
\end{table}

\begin{table}[t]

    \begin{center}
    \setlength{\tabcolsep}{1pt}
    \begin{small}
    \begin{tabular}{c|ccc|cccccccc}
    \hline
    \textbf{Method} & \textbf{PromptEOL} & \textbf{PromptEOL+TP}      \\
    \hline
    
    SprintDuplicateQuestions  & 43.02                         & 51.61\\
    TwitterSemEval2015        & 65.61                         & 67.70\\
    TwitterURLCorpus          & 78.97                         & 80.90\\
     
    \hline
    \bf{Average (3)}         & 62.53                         & 66.74\\
    \hline
    \end{tabular}
    \caption{Results (accuracy scaled by 100x) on pair classification datasets using LLaMA2-7B.}
    \label{table:mteb_pair_classification}
    \end{small}
    \end{center}
\end{table}

\begin{table}[th]

    \begin{center}
    \setlength{\tabcolsep}{1pt}
    \begin{small}
    \begin{tabular}{c|ccc|cccccccc}
    \hline
    \textbf{Method} & \textbf{PromptEOL} & \textbf{PromptEOL+TP}      \\
    \hline
    
    AskUbuntuDupQuestions     & 53.88                         & 57.02\\
    MindSmallRerank           & 29.97                         & 29.89\\
    SciDocsRR                 & 71.38                         & 77.49\\
    StackOverflowDupQuestions & 40.63                         & 43.19\\ 
     
    \hline
    \bf{Average (4)}         & 48.97                         & 51.90\\
    \hline
    \end{tabular}
    \caption{Results (average precision scaled by 100x) on reranking datasets using LLaMA2-7B.}
    \label{table:mteb_reranking}
    \end{small}
    \end{center}
\end{table}

The datasets we used are all from the MTEB benchmark \cite{muennighoff2022mteb}.
The 12 classification datasets include AmazonCounterfactual, AmazonPolarity, AmazonReviews, Banking77, Emotion, Imdb, MassiveIntent, MassiveScenario, MTOPDomain, MTOPIntent, ToxicConversations, and TweetSentimentExtraction. 
The 3 pair classification datasets are SprintDuplicateQuestions, TwitterSemEval2015, and TwitterURLCorpus. 
The 4 reranking datasets are AskUbuntuDupQuestions, MindSmallRerank, SciDocsRR, and StackOverflowDupQuestions.
The 11 clustering datasets are  ArxivClusteringP2P,ArxivClusteringS2S, BiorxivClusteringP2P, BiorxivClusteringS2S, MedrxivClusteringP2P, MedrxivClusteringS2S, RedditClustering, RedditClusteringP2P, StackExchangeClustering, StackExchangeClusteringP2P, TwentyNewsgroupsClustering.
The summarization dataset is SummEval.
The additional STS dataset is BIOSSES.

The results on classification datasets, pair classification datasets, reranking datasets, clustering datasets, retrieval datasets, summarization dataset and additional STS dataset are presented in Table \ref{table:mteb_classification}, Table \ref{table:mteb_pair_classification}, Table \ref{table:mteb_reranking}, Table \ref{table:mteb_clustering}, Table \ref{table:mteb_summarization}, and Table \ref{table:mteb_sts}, respectively. 
Our method shows improvement in 40 out of 44 cases across all datasets. Specifically, the TP technique achieves an average improvement of 2.72 on classification datasets, a 4.21 increase on pair classification datasets, a 2.93 gain on reranking datasets, an improvement of 4.51 on clustering datasets, a 1.23 gain on summarization dataset, and a 7.07 gain on additional STS dataset.

\begin{table}[t]

    \begin{center}
    \setlength{\tabcolsep}{1pt}
    \begin{small}
    \begin{tabular}{c|ccc|cccccccc}
    \hline
    \textbf{Method} & \textbf{PromptEOL} & \textbf{PromptEOL+TP}      \\
    \hline
    
    ArxivClusteringP2P         &34.87                        &43.57 \\
    ArxivClusteringS2S         &31.19                        &39.82 \\
    BiorxivClusteringP2P       &19.56                        &25.75 \\
    BiorxivClusteringS2S       &24.34                        &31.92 \\ 
    MedrxivClusteringP2P       &27.65                        &26.41 \\
    MedrxivClusteringS2S       &34.53                        &35.98 \\
    RedditClustering           &24.69                        &31.69 \\
    RedditClusteringP2P        &48.52                        &48.54 \\
    StackExchangeClustering    &42.16                        &44.90 \\
    StackExchangeClusteringP2P &33.56                        &33.03 \\
    TwentyNewsgroupsClustering &27.61                        &36.98 \\
    
    \hline
    \bf{Average (11)}          &31.70                        &36.24 \\
    \hline
    \end{tabular}
    \caption{Results (V-measure scaled by 100x) on clustering datasets using LLaMA2-7B.}
    \label{table:mteb_clustering}
    \end{small}
    \end{center}
\end{table}

\begin{table}[th]
    
    \begin{center}
    \setlength{\tabcolsep}{10pt}
    \begin{small}
    \begin{tabular}{c|ccc|cccccccc}
    \hline
    \textbf{Method} & \textbf{PromptEOL} & \textbf{PromptEOL+TP}      \\
    \hline

    SummEval  & 28.88                         & 30.11 \\

    \hline
    \end{tabular}
    \caption{Results (Spearman correlation scaled by 100x) on summarization dataset using LLaMA2-7B.}
    \label{table:mteb_summarization}
    \end{small}
    \end{center}
\end{table}

\begin{table}[t]
    
    \begin{center}
    \setlength{\tabcolsep}{10pt}
    \begin{small}
    \begin{tabular}{c|ccc|cccccccc}
    \hline
    \textbf{Method} & \textbf{PromptEOL} & \textbf{PromptEOL+TP}      \\
    \hline

    BIOSSES  & 62.66                         & 69.73 \\

    \hline
    \end{tabular}
    \caption{Results (Spearman correlation scaled by 100x) on additional STS dataset using LLaMA2-7B.}
    \label{table:mteb_sts}
    \end{small}
    \end{center}
\end{table}

\subsection{More Prompt Baseline Evaluation}
\label{sec:appendix_prompt}
We identify two prompts A and B similar to PromptEOL that could benefit more significantly from TP. These prompts are derived from \cite{li2024bellm} and \cite{li2023angle}. In addition, we design two prompts C and D to impart clear semantic information to the {\textless \text{PST}\textgreater} token. The specific prompts are shown below:

\begin{tcolorbox}
\textbf{Prompt A:}  
"The representative word for sentence <PST> '[TEXT]' is:"

\textbf{Prompt B:}  
"Summarize sentence <PST> '[TEXT]' in one word:"

\textbf{Prompt C:}  
"Given the keyword <PST>, this sentence: '[TEXT]' means in one word:"

\textbf{Prompt D:}  
"This sentence: <PST> and '[TEXT]' means in one word:"

\end{tcolorbox}

We conduct comparative experiments with and without TP using Prompt A and B. The results are shown in the Table \ref{tab:ab_prompt}.
As shown in the table, our proposed method significantly improves the performance of  prompt A and B, achieving a 10.64 and 9.26 increase, respectively. This validates our hypothesis that simple prompts without prior knowledge, similar to PromptEOL, rely more heavily on modeling backward dependencies to effectively capture semantics.

We observe that compared to the results in Table \ref{tab:results}, Prompt C and D do not further enhance TP’s performance in the Table \ref{tab:ab_prompt}. We speculate this is because TP edits occur in the intermediate layers of the LLM, and providing prior knowledge about the <PST> token in the input does not effectively help the model grasp its intended meaning.

\begin{table*}[t!]
\centering
\normalsize
\setlength{\tabcolsep}{5pt}
\resizebox{\linewidth}{!}{%
\begin{tabular}{lrccccccccc}
\toprule
\textbf{Method} & \textbf{Backbone} & \textbf{STS12} & \textbf{STS13} & \textbf{STS14} & \textbf{STS15} & \textbf{STS16} & \textbf{STS-B} & \textbf{SICK-R} & \textbf{Avg.}\\
\midrule
\midrule
Prompt A & LLaMA2-7B & 44.79&	65.73&	50.39&	58.70&	58.10&	51.42&	47.92&	53.86  \\
Prompt A + TP (\textbf{\textit{Ours}}) & LLaMA2-7B & \cellcolor{gray!20}52.22& \cellcolor{gray!20}67.44& \cellcolor{gray!20}58.00& \cellcolor{gray!20}69.89& \cellcolor{gray!20}71.32& \cellcolor{gray!20}65.76& \cellcolor{gray!20}66.86& \cellcolor{gray!20}64.50\\

\midrule
Prompt B & LLaMA2-7B & 51.18&	73.74&	63.13&	68.87&	70.96&	63.29&	67.45&	65.52 \\
Prompt B + TP (\textbf{\textit{Ours}}) & LLaMA2-7B & \cellcolor{gray!20}64.32& \cellcolor{gray!20}80.18& \cellcolor{gray!20}70.49& \cellcolor{gray!20}77.29& \cellcolor{gray!20}78.36& \cellcolor{gray!20}79.32& \cellcolor{gray!20}73.47& \cellcolor{gray!20}74.78\\
\midrule
\midrule
Prompt C & LLaMA2-7B & 64.34&	77.87&	67.62&	74.25&	72.15&	77.33&	74.91&	72.64 \\
\midrule
Prompt D & LLaMA2-7B & 61.99&	80.83&	71.69&	78.31&	77.06&	77.82&	73.68&	74.48 \\
\bottomrule
\end{tabular}
}
\caption{Results on STS tasks (Spearman correlation scaled by 100x) using prompt A, B, C, and, D in Appendix \ref{sec:appendix_prompt}.}
\label{tab:ab_prompt}
\end{table*}

\begin{table*}[t!]
\centering
\normalsize
\setlength{\tabcolsep}{5pt}
\resizebox{\linewidth}{!}{%
\begin{tabular}{lrccccccccc}
\toprule
\textbf{Method} & \textbf{Backbone} & \textbf{STS12} & \textbf{STS13} & \textbf{STS14} & \textbf{STS15} & \textbf{STS16} & \textbf{STS-B} & \textbf{SICK-R} & \textbf{Avg.}\\
\midrule
\midrule
1 <PST> (PromptEOL+TP)& LLaMA2-7B & 66.90&	83.12&	74.31&	79.87&	80.03&	80.67&	75.40&	77.19
 \\
\midrule
2 <PST> & LLaMA2-7B & 65.98&	83.34&	74.31&	80.31&	80.17&	80.78&	76.01&	77.27 \\
\midrule
3 <PST> & LLaMA2-7B & 64.98&	82.92&	72.99&	79.33&	79.34&	80.01&	75.95&	76.50 \\
\midrule
4 <PST> & LLaMA2-7B & 64.52&	82.90&	72.01&	78.90&	78.67&	79.50&	75.75&	76.04 \\
\bottomrule
\end{tabular}
}
\caption{Results on STS tasks (Spearman correlation scaled by 100x) based on PromptEOL, with various number of <PST> placeholder tokens incorporated into the prompt.}
\label{tab:pst_num}
\end{table*}

\begin{table*}[t!]
\centering
\normalsize
\setlength{\tabcolsep}{5pt}
\resizebox{\linewidth}{!}{%
\begin{tabular}{lrccccccccc}
\toprule
\textbf{Method} & \textbf{Backbone} & \textbf{STS12} & \textbf{STS13} & \textbf{STS14} & \textbf{STS15} & \textbf{STS16} & \textbf{STS-B} & \textbf{SICK-R} & \textbf{Avg.}\\
\midrule
\midrule
PromptEOL+TP & LLaMA2-7B & 66.90&	83.12&	74.31&	79.87&	80.03&	80.67&	75.40&	77.19
 \\
\midrule
Masking <PST> & LLaMA2-7B & 66.85&	82.97&	74.17&	79.72&	79.94&	80.55&	75.26&	77.07 \\

\bottomrule
\end{tabular}
}
\caption{Results on STS tasks (Spearman correlation scaled by 100x) based on PromptEOL, masking the <PST> token in the first layer.}
\label{tab:pst_mask}
\end{table*}

\begin{table*}[t!]
\centering
\normalsize
\setlength{\tabcolsep}{5pt}
\resizebox{\linewidth}{!}{%
\begin{tabular}{lrccccccccc}
\toprule
\textbf{Method} & \textbf{Backbone} & \textbf{STS12} & \textbf{STS13} & \textbf{STS14} & \textbf{STS15} & \textbf{STS16} & \textbf{STS-B} & \textbf{SICK-R} & \textbf{Avg.}\\
\midrule
\midrule
PromptEOL+TP & LLaMA2-7B & 66.90&	83.12&	74.31&	79.87&	80.03&	80.67&	75.40&	77.19
 \\
\midrule
Resuming at Layer 9 & LLaMA2-7B & 66.97	&82.75	&73.24	&78.75	&79.36	&80.40	&75.33	&76.69 \\
Resuming at Layer 10 & LLaMA2-7B & 66.76	&82.91	&73.26	&79.11	&79.40	&80.30	&75.58	&76.76 \\
Resuming at Layer 11 & LLaMA2-7B & 66.68	&83.01	&73.34	&79.13	&79.46	&80.17	&75.60	&76.77 \\
Resuming at Layer 16 & LLaMA2-7B & 66.84	&83.07	&74.22	&79.83	&79.96	&80.56	&75.36	&77.12 \\
Resuming at Layer 26 & LLaMA2-7B & 66.95	&83.18	&74.33	&79.89	&80.04	&80.66	&75.40	&77.21
\\
\bottomrule
\end{tabular}
}
\caption{Results on STS tasks (Spearman correlation scaled by 100x) based on PromptEOL, resuming TP at different layer.}
\label{tab:resuming_tp}
\end{table*}

\begin{table*}[t!]
\centering
\normalsize
\setlength{\tabcolsep}{5pt}
\resizebox{\linewidth}{!}{%
\begin{tabular}{lrccccccccc}
\toprule
\textbf{Method} & \textbf{Backbone} & \textbf{STS12} & \textbf{STS13} & \textbf{STS14} & \textbf{STS15} & \textbf{STS16} & \textbf{STS-B} & \textbf{SICK-R} & \textbf{Avg.}\\
\midrule
\midrule
w/o prompt & LLaMA2-7B & 9.13&	22.25&	11.04&	33.09	&34.92	&15.94	&33.73	&22.87
 \\
\midrule
w/o prompt + TP & LLaMA2-7B & 11.68	&25.85	&13.08	&33.70	&46.58	&22.05	&42.61	&27.94
 \\

\bottomrule
\end{tabular}
}
\caption{Results on STS tasks (Spearman correlation scaled by 100x) based on PromptEOL without any prompt.}
\label{tab:wo_prompt}
\end{table*}

\subsection{Number of {\textless \text{PST}\textgreater} tokens}

We further analyze the impact of the number of inserted {\textless \text{PST}\textgreater} tokens on performance based on PromptEOL. The results are shown in the Table\ref{tab:pst_num}.
Incorporating two <PST> tokens achieve a slight improvement in TP performance (by 0.08 points). However, prepending more <PST> tokens leads to a decline in performance, as evidenced by the results of 3 {\textless \text{PST}\textgreater} tokens and 4 {\textless \text{PST}\textgreater} tokens.

\subsection{Masking <PST> Token in the First Layer}
We mask the <PST> token in the first layer of the LLaMA2 7B model to mitigate the impact of token initialization. The experimental results are shown in the Table \ref{tab:pst_mask}.

The performance is slightly lower than that of PromptEOL+TP. This may be attributed to the role of the <PST> token in the first layer, where it acts as a placeholder, enabling the LLM to interpret the input length as N+1. Despite its random initialization, the <PST> token ensures consistent input length across all layers.

\subsection{Resuming TP at Different Layers}
We explore the practice of using TP for a few layers, and then pause for a few layers and then resume again. We conduct experiments based on the best practices of PromptEOL and TP, applying TP at the 1st layer, stopping TP at the 8th layer, and early exiting at the 27th layer. The results are shown in the Table \ref{tab:resuming_tp}. Once resumed, TP remains active until output. We find that resuming TP at deeper layers (e.g., layer 21) yields a slight performance improvement.

\subsection{TP without Prompt}
We test performance of TP without any prompt. The results are shown in the Table\ref{tab:wo_prompt}. Although the absence of a prompt significantly degrades performance, TP still manages to provide improvements.

\end{document}